\documentclass[preprint,12pt]{elsarticle}

\usepackage{amsmath,amssymb,amsfonts}
\usepackage{mathtools}
\usepackage{bm}
\usepackage{siunitx}

\usepackage{booktabs}
\usepackage{array}
\usepackage{tabularx}
\usepackage{multirow}
\usepackage{ragged2e}
\newcolumntype{L}{>{\RaggedRight\arraybackslash}X}
\usepackage{adjustbox}

\usepackage{subcaption}
\usepackage{enumitem}

\usepackage{tikz}
\usetikzlibrary{positioning,arrows.meta,fit,shapes.geometric,calc}

\usepackage[hyphens]{url}

\biboptions{sort&compress}
\journal{Computers in Biology and Medicine}

\begin{document}

\begin{frontmatter}

\title{Recognition and Label-Free Adaptation Across Recording Sessions in Surface-EMG
Gesture Decoding}

\author[bme]{Jethro Odeyemi}
\ead{jethro.odeyemi@usask.ca}

\author[bme]{W.J. (Chris) Zhang\corref{cor1}}
\ead{chris.zhang@usask.ca}
\cortext[cor1]{Corresponding author.}

\affiliation[bme]{organization={Division of Biomedical Engineering, University of Saskatchewan},
            addressline={57 Campus Drive},
            city={Saskatoon},
            state={SK},
            postcode={S7N 5A9},
            country={Canada}}

\begin{abstract}
Recognition accuracy obtained during a recording session does not persist when a user puts on
the electrodes again after the electrodes had previously been removed. The electrodes may have
moved slightly, the skin may be drier or wetter, or the elbow may be positioned differently;
these factors all contribute to day-to-day variability and therefore represent a major obstacle
to implementing successful pattern-recognition based myoelectric control systems in daily
practice. However, simply recalibrating a user's hand for 20 min at every doff/don
event is a clearly unrealistic expectation. A montage-agnostic encoder built for cross-user,
cross-montage transfer is trained here using data collected during a particular recording
session, and then applied to data collected later in a different recording session without
adjusting anything, on the ten intact subjects of NinaPro DB6. The performance of this approach
is compared to that of a per-user LDA classification pipeline, and to that of two published
approaches that only rely on source data collected from the same recording session. Carried
unchanged across recording sessions, the encoder retains 0.688 macro-F1 against 0.540 for the
per-user pipeline, and, on the per-window metric the published baselines use, sits above both
published source-only results, a band of two points that locates the encoder rather than
ranking it. Of five label-free test-time
adaptations, only feature-statistic alignment improves every subject; batch-normalisation
re-estimation, a standard method in the domain-adaptation literature, collapses this
architecture entirely. Aligning the encoder's feature statistics to the new session recovers
about what a single labelled calibration repetition would.
\end{abstract}


\end{frontmatter}

\section{Introduction}

Recognition accuracy obtained during a recording session does not persist when a user puts on
the electrodes again after the electrodes had previously been removed. There are several reasons
why there will be changes between recordings. For example, the electrodes may have moved
slightly due to changes in the way the electrodes sit against the user's body; the user's skin
may be drier or wetter than before; or the user's elbow may be positioned differently than
before. In addition, the muscle activation patterns that produce signals that can be separated
by a classifier may fall outside of the regions of muscle space that the classifier has been
trained to recognize. These factors all contribute to day-to-day variability and therefore
represent a major obstacle to implementing successful pattern-recognition based myoelectric
control systems in daily
practice~\citep{palermo2017repeatability,du2017surface,zhai2017self,jiang2017feasibility}.
However, simply recalibrating a user's hand for 20 min at every doff/don event is a clearly
unrealistic expectation.

This paper builds upon a montage-agnostic encoder developed in earlier
work~\citep{odeyemi_encoder}. That encoder successfully transferred between users and between
different electrode montages without requiring individualized calibration. However, that work
assumed that the recording session remained constant. Specifically, training and testing
occurred within the same recording session. Therefore, the primary objective of this paper is to
measure how much of this recognition remains valid when the recording session is different from
the training session. In other words, given that the electrodes have been taken off and put back
on, how much of the recognition established in one day will remain valid on subsequent days? How
can we address that portion of recognition that does not remain valid without having the user
provide additional labeled information?

Therefore, this paper describes two sets of experiments. The first set concerns the encoder
itself. The same montage-agnostic model described in Section~\ref{sec:p3-encoder} is trained
using data collected during a particular recording session, and then applied to data collected
later in a different recording session without adjusting anything. The performance of this
approach is compared to that of a per-user LDA classification pipeline. Additionally, the
performance of the encoder is compared to that of two published approaches that only rely on
source data collected from the same recording session. The second set concerns adapting to a new
recording session for which labeled data is not available. Although there are many adaptation
methods that have recently been introduced in the literature to adapt a pre-trained model to a
new task (to be discussed), these methods do not perform uniformly on this task. The only
label-free adaptation method among these that improves performance for all subjects is aligning
the pooled feature statistics of the encoder to those of the new session.

\section{Background and positioning}

As mentioned earlier, degradations in surface EMG recordings between sessions have been long
known and widely documented in the literature. As such, despite many advances in machine
learning techniques for analyzing EMG signals, current clinical pipelines still rely heavily on
traditional time-domain feature extraction~\citep{hudgins1993new} followed by per-user linear
discriminant analysis (LDA) classifiers. However, even these traditional pipelines have been
shown to degrade significantly with respect to electrode movement and time since last
recording~\citep{palermo2017repeatability}. The most common solution currently implemented
clinically is to recalculate the classifier each time a patient returns for a follow-up visit.
Another strategy is to collect training data from multiple visits (or sessions), allowing the
classifier to "average" across some degree of
variability~\citep{du2017surface,zhai2017self,jiang2017feasibility,rehman2018multiday,qi2019intelligent}.
While neither strategy requires significant computational resources or expertise, both require
significant input from the clinician/patient.

In contrast, researchers have framed this problem as domain adaptation, where the source
corresponds to the training session, and the target corresponds to the new/evaluation session.
Several key concepts are relevant here. Adversarial alignment learns features that a domain
discriminator cannot distinguish~\citep{ganin2016domain}; this technique has been extended to
include deep neural networks for inter-session EMG data~\citep{du2017intersession} and recurrent
EMG models~\citep{ketyko2019domain}. Statistical alignment provides a less computationally
expensive approach, matching the covariance~\citep{sun2016deepcoral}, the second-order
statistical properties of the feature distributions in both domains. Batch normalization
re-estimation is another statistical alignment approach that leaves the weight matrices
unchanged while estimating new batch normalization statistics on the target
domain~\citep{li2018adabn}. Test-time entropy-based adaptation updates a small subset of
parameters to improve confidence in the model's predictions on unlabeled target
data~\citep{wang2021tent}. Confidence-filtered pseudo-labelling uses those predicted values as
pseudo-target labels~\citep{lee2013pseudo}. Finally, convolutional and temporal-convolutional
models serve as supervised reference points on
DB6~\citep{zanghieri2020temponet,lin2023longterm}, often incorporating some level of
target-session data into training.

The model presented in this paper is the same one described in
Section~\ref{sec:p3-encoder}~\citep{odeyemi_encoder}, to serve as a baseline for measuring how
robust a model for montage-invariant recognition is when tested against session-variability.
Furthermore, since each of these test-time adaptation methods will be treated as probes, rather
than contributions themselves, this paper will report on how each performs on this architecture
and data.

\section{Method}

\subsection{Encoder architecture}
\label{sec:p3-encoder}

The model is identical to the montage-agnostic encoder introduced for calibration-light
cross-user gesture recognition~\citep{odeyemi_encoder}: causal rolling-time normalization per
channel, a shared per-channel convolutional tokenizer, a learned electrode coordinate positional
encoding, a cross-channel transformer, and attention pooling into a linear head (about 3.3M
parameters total). Since nothing in this architecture has changed for the inter-session case,
what is being measured is whether or not a model originally optimized for recognizing a
different type of transfer exhibits session-robustness rather than whether or not a model
specifically engineered for this type of transfer exhibits session-robustness.

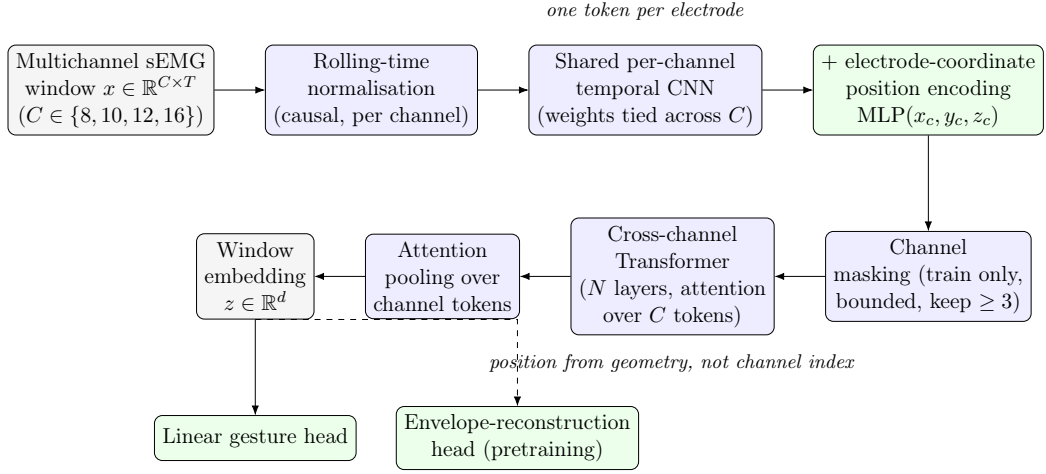
\begin{figure}[t]
\centering
\resizebox{\textwidth}{!}{%
\begin{tikzpicture}[
  font=\small,
  box/.style={draw, rounded corners, align=center, minimum height=8mm, inner sep=4pt, fill=black!4},
  op/.style={draw, rounded corners, align=center, minimum height=8mm, inner sep=4pt, fill=blue!7},
  hl/.style={draw, rounded corners, align=center, minimum height=8mm, inner sep=4pt, fill=green!8},
  ->, >=Latex, node distance=6mm and 9mm]

\node[box] (in) {Multichannel sEMG\\window $x\in\mathbb{R}^{C\times T}$\\($C\in\{8,10,12,16\}$)};
\node[op, right=of in] (rtn) {Rolling-time\\normalisation\\(causal, per channel)};
\node[op, right=of rtn] (cnn) {Shared per-channel\\temporal CNN\\(weights tied across $C$)};
\node[hl, right=of cnn] (pos) {$+$ electrode-coordinate\\position encoding\\$\mathrm{MLP}(x_c,y_c,z_c)$};

\node[op, below=17mm of pos] (acm) {Channel\\masking (train only,\\bounded, keep $\ge 3$)};
\node[op, left=of acm] (tf) {Cross-channel\\Transformer\\($N$ layers, attention\\over $C$ tokens)};
\node[op, left=of tf] (pool) {Attention\\pooling over\\channel tokens};
\node[box, left=of pool] (out) {Window\\embedding\\$z\in\mathbb{R}^{d}$};

\draw (in) -- (rtn);
\draw (rtn) -- (cnn);
\draw (cnn) -- (pos);
\draw (pos.south) -- (acm.north);
\draw (acm) -- (tf);
\draw (tf) -- (pool);
\draw (pool) -- (out);

\node[align=center, font=\footnotesize, above=3mm of cnn] {\emph{one token per electrode}};
\node[align=center, font=\footnotesize, fill=white, inner sep=1pt, below=3mm of tf]
  {\emph{position from geometry, not channel index}};

\node[hl, below=17mm of out] (head) {Linear gesture head};
\node[hl, right=7mm of head] (mae) {Envelope-reconstruction\\head (pretraining)};
\draw (out.south) -- (head.north);
\draw[dashed] (out.south) -| (mae.north);
\end{tikzpicture}%
}
\caption[Fused montage-agnostic encoder]{The proposed montage-agnostic encoder. Each electrode
becomes a single token whose position is supplied by its forearm coordinate rather than its channel
index, so one set of weights ingests any electrode count. Rolling-time normalisation makes the
representation calibration-free, cross-channel attention mixes the electrode tokens, and bounded
channel masking regularises toward user-invariant features. The linear head is used for gesture
classification; the dashed head is the self-supervised pretraining branch.}
\label{fig:ch3-encoder}
\end{figure}

We begin by providing details concerning our encoder's design. The overall architecture is
depicted in Figure~\ref{fig:ch3-encoder}. Our encoder maps sequences of EMG signals ($x \in
\mathbb{R}^{C\times T}$) onto fixed-size embeddings ($z$), where $C$ represents the possible
number of channels in any of the available montages.

Our proposed encoder is founded on the idea that each electrode is treated as one independent
token whose identity corresponds to its physical position on the forearm.

\paragraph{Causal Rolling-Time normalisation.} Our first layer normalizes each channel
individually using their respective causal rolling statistics over time; these statistics
represent expanding window means and variances calculated over time. Since our normalizing is
done per channel and only uses causal information from past time steps, at runtime a previously
unobserved user will self-normalize their input data with respect to their own signal using only
their own causal history; no user-specific statistics will be transported from training to
testing. Hence, even though there is no labeled data from the new user available at runtime, our
model will still adaptively scale its input representations according to new users' signals
without requiring any additional labels.

\paragraph{Shared per-channel tokeniser.} To produce one feature vector per electrode, we apply
a single one-dimensional convolutional stack separately to every channel. Due to sharing of
weights across channels, adding or removing an electrode simply adds/removes a token to/from our
tokenizer without modifying any parameter(s).

\paragraph{Electrode-coordinate position encoding.} Each electrode contains normalized $(x,y,z)$
coordinates specifying its position on the forearm. These coordinates are mapped to vectors
added to each electrode token by means of a multilayer perceptron. Therefore, our model learns
where each electrode resides, not merely its index, thus allowing any electrode configuration to
map into one uniform representation space, with each electrode serving as an unordered element
within that space.

\paragraph{Cross-channel attention.} Our transformer encoder~\citep{vaswani2017attention}
applies attention mechanisms across our electrode tokens. The temporal structure is encoded via
our per-channel tokenizers; attention mechanisms capture spatial relations between electrodes.
Following application of attention mechanisms across all channels, an attention-pooling layer is
used to collapse the electrode tokens into a single window embedding; finally a linear head
computes gesture logits.

\paragraph{Channel-masking regularisation.} During training time, a bounded number of electrodes
is randomly masked such that at least three channels remain active for each montage. This forces
our model towards features that are invariant or robust with regards to missing or displaced
electrodes rather than relying on any single channel. How many electrodes should be masked
affects whether masking fails completely: if too aggressive on very small montages (e.g., all
electrodes masked), there are no valid windows containing active electrodes resulting in
collapsed training; we solved this issue by enforcing bounds for masking.

\subsection{Inter-session protocol}

\begin{figure}[tbp]\centering
\begin{subfigure}{\textwidth}\centering
\resizebox{\textwidth}{!}{%
\begin{tikzpicture}[
  font=\small,
  box/.style={rounded corners=2pt, draw, minimum height=8mm, inner xsep=6pt, align=center},
  sess/.style={box, fill=black!4},
  train/.style={box, fill=blue!8, draw=blue!55},
  test/.style={box, fill=red!7, draw=red!55},
  cal/.style={box, fill=green!9, draw=green!55},
  arr/.style={-{Latex[length=2mm]}, thick},
  lbl/.style={font=\footnotesize\itshape, text=black!60, fill=white, inner sep=1.5pt}]

\node[sess, minimum width=52mm] (a) at (0,2.2) {Session a \; (day 1)};
\node[train, minimum width=34mm] (atr) at (-0.9,1.2) {train repetitions\\ \footnotesize (reps not divisible by 5)};
\node[test, minimum width=14mm] (ate) at (1.85,1.2) {held-out\\ \footnotesize reps};

\node[sess, minimum width=52mm] (b) at (0,-0.9) {Session b \; (later day)};
\node[test, minimum width=50mm] (bte) at (0,-1.9) {all repetitions \; (unseen session)};

\node[box, fill=black!8, minimum width=30mm, minimum height=11mm] (m) at (7.2,0.7)
  {Montage-agnostic\\ encoder\\ \footnotesize trained on session a};

\draw[arr, blue!60] (atr.east) .. controls +(1.2,0) and +(-1.2,0) .. (m.west);
\draw[arr, red!55] (ate) .. controls +(2.4,0) and +(-1.4,0.6) .. node[lbl, above, pos=0.6]{within-session eval} (m.170);
\draw[arr, red!60] (bte.east) .. controls +(2.6,0) and +(-1.4,-0.6) .. node[lbl, below, pos=0.6]{cross-session eval} (m.190);

\node[cal, minimum width=30mm, align=center] (adapt) at (7.2,-1.9)
  {label-free alignment\\ \footnotesize or 1--3 shot calibration};
\draw[arr, green!55] (m.south) -- node[lbl, right]{adapt on session b} (adapt.north);
\draw[arr, green!55] (adapt.west) .. controls +(-1.4,0) and +(2.6,-0.2) .. (bte.east);

\end{tikzpicture}%
}
\caption{}\label{fig:ch4-protocol}
\end{subfigure}

\vspace{4mm}
\begin{subfigure}{\textwidth}\centering
\resizebox{\textwidth}{!}{%
\begin{tikzpicture}[
  font=\small,
  b/.style={rounded corners=2pt, draw, minimum height=8mm, inner xsep=5pt, align=center},
  src/.style={b, fill=blue!8, draw=blue!55},
  tgt/.style={b, fill=red!7, draw=red!55},
  op/.style={b, fill=green!10, draw=green!55},
  bad/.style={b, fill=black!6, draw=black!45},
  arr/.style={-{Latex[length=2mm]}, thick}]

\node[align=right, anchor=east, font=\footnotesize\bfseries] at (-1.75,2.2) {Feature alignment\\ (label-free)};
\node[src, minimum width=20mm] (s1) at (0,2.2) {source stats\\ \footnotesize $\mu_s,\sigma_s$ (session a)};
\node[tgt, minimum width=20mm] (t1) at (3.7,2.2) {target stats\\ \footnotesize $\mu_t,\sigma_t$ (session b)};
\node[op, minimum width=22mm] (o1) at (7.6,2.2) {align embedding\\ \footnotesize $z \gets \sigma_s\frac{z-\mu_t}{\sigma_t}+\mu_s$};
\draw[arr, red!55] (t1) -- (o1);
\draw[arr, blue!55] (s1.north) .. controls +(0,0.7) and +(0,0.7) .. (o1.north);
\node[b, fill=green!16, minimum width=16mm] (h1) at (11.0,2.2) {head\\ \footnotesize (frozen)};
\draw[arr] (o1) -- (h1);
\node[font=\footnotesize, text=green!45!black] at (11.0,1.4) {$+0.029$, all ten subjects};

\node[align=right, anchor=east, font=\footnotesize\bfseries] at (-1.75,-0.7) {BatchNorm re-estimation\\ (AdaBN)};
\node[bad, minimum width=20mm] (c1) at (0,-0.7) {tokenizer BN\\ \footnotesize per-channel conv};
\node[bad, minimum width=26mm] (c2) at (4.4,-0.7) {re-estimate BN on\\ \footnotesize channel-mixed target batch};
\node[b, fill=red!12, draw=red!60, minimum width=22mm] (c3) at (9.0,-0.7) {corrupted statistics};
\draw[arr] (c1) -- (c2);
\draw[arr, red!60] (c2) -- (c3);
\node[b, fill=red!22, draw=red!70, minimum width=18mm] (c4) at (12.4,-0.7) {one class\\ \footnotesize (collapse)};
\draw[arr, red!70] (c3) -- (c4);
\node[font=\footnotesize, text=red!60!black] at (9.0,-1.6) {$-0.641$, every subject};

\end{tikzpicture}%
}
\caption{}\label{fig:ch4-adaptation}
\end{subfigure}
\caption{Experimental design. (a) Inter-session protocol on DB6. The encoder trains only on
session a. It is scored within session a on held-out repetitions and, without retraining,
across the day gap on the whole of session b. Label-free alignment and light calibration act
on session b alone, never on its labels for the label-free path. (b) Two ways to adapt to the
new session. Aligning the pooled embedding's first two moments to the source keeps the frozen
head calibrated and helps every subject. Re-estimating BatchNorm statistics inside the
per-channel tokenizer, where each batch mixes electrodes, corrupts the running statistics and
collapses the model to a single class.}
\label{fig:ch4-method}
\end{figure}
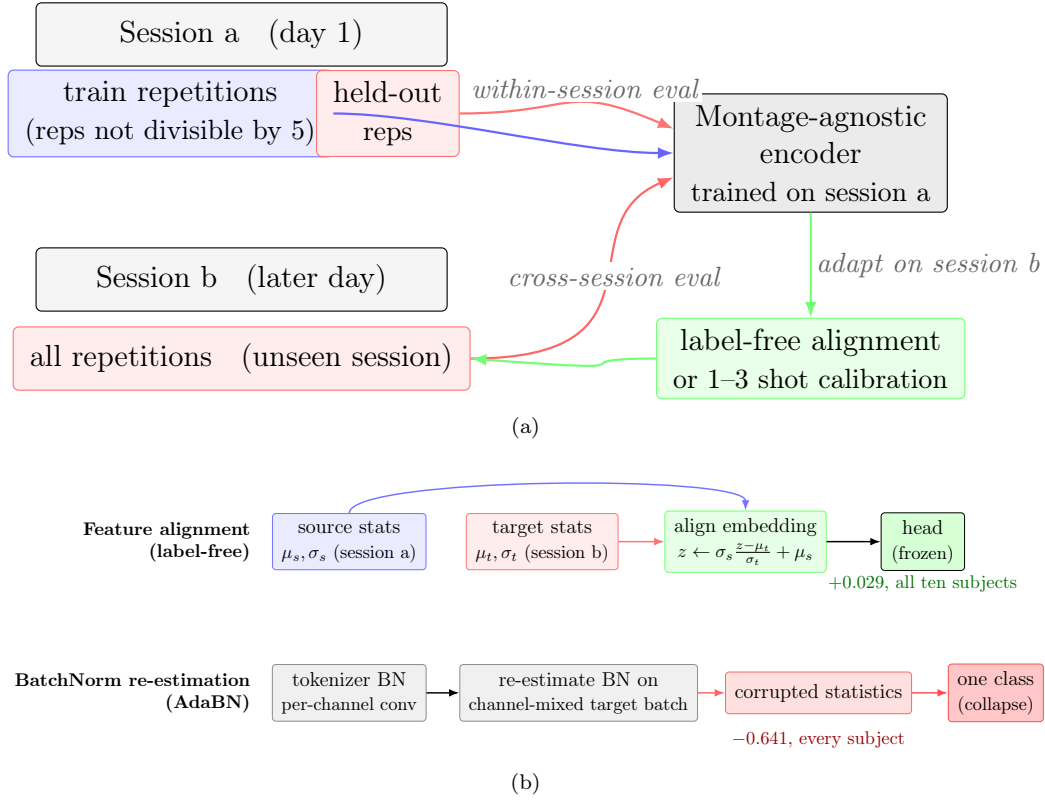

Fig.~\ref{fig:ch4-protocol} shows a high-level overview of our experimental design. Each of the
ten subjects included in DB6 contributed two separate sessions recorded on different days with
their electrodes removed and reapplied. We trained our encoder on session A, using the
repetitions whose index is not divisible by five, and evaluated its performance in two regimes.
In the within-session regime, we evaluate its performance using only the held-out repetitions
from Session A. Therefore, we are measuring only ordinary generalization. In the cross-session
regime, we use all of Session B. Thus, we are evaluating what aspects of its performance survive
a day gap. Our encoder was never re-trained between evaluations.

\subsection{Label-free adaptation}

When a new Session B arrives with no labels, adaptation methods can only utilize unlabeled
target windows. Therefore, we investigate five possible approaches that operate solely on
Session B.

\emph{Feature alignment} rescales the pooled representation of Session B so that its
per-dimensional means and variances match those measured on Session A. Then we apply our frozen
head (See Fig.~\ref{fig:ch4-adaptation}, top). Feature Alignment assumes that the boundaries
defined by our classifier are correct and that differences between sessions occur mainly as an
affine transformation of our feature distributions.

\emph{Entropy minimisation} utilizes Wang et al.'s entropy-based test-time adaptation
method~\citep{wang2021tent} and adjusts our model to increase its confidence in its own
predictions on the target domain. By itself this collapses to predicting just one class; thus,
we add a marginal-entropy regularizer to maintain some spread in our predicted class
distribution.

\emph{Pseudo-label self-training}~\citep{lee2013pseudo} uses confident target predictions as
pseudo-labels.

\emph{Batch-normalisation re-estimation}~\citep{li2018adabn} estimates new batch normalization
statistics on Session B, replacing those estimated on Session A. We experiment with two
locations for batch normalization layers: one location near the end of our feature hierarchy
(just prior to applying our frozen head), and one location deeper in our feature hierarchy
(throughout our per-channel tokenizer) (See Fig.~\ref{fig:ch4-adaptation}, bottom).


We also include an additional light-calibration path in our results in order to compare how
effective each of our label-free adaptation methods are versus simply fine-tuning on one or
three labeled repetitions from Session B.

\subsection{Metrics}

Throughout this paper we will report two metrics. Trial-Voted Macro-F1 assigns one prediction
per gesture repetition by majority vote over its windows and averages F1 across classes. We
chose Macro-F1 because it is robust to class imbalance; it was used in the cross-user
evaluation of this encoder~\citep{odeyemi_encoder}. Per-window accuracy scores every window
independently and was chosen because it matches how prior work reported accuracy numbers. Since
both metrics allow us to compare to prior work fairly (Macro-F1 represents decisions regarding
gestures whereas per-window accuracy represents decisions made per window), we chose to report
both metrics.

\section{Experimental setup}

\begin{table}[tbp]\centering
\caption{Experimental setup. (a) DB6 inter-session benchmark. Ten intact subjects each
recorded two sessions (a, b) on different days; the model trains on session a and is
evaluated on the unseen session b. (b) Training configuration for the DB6 inter-session
experiments.}\label{tbl:ch4-setup}
\begin{subtable}{\textwidth}\centering
\caption{}\label{tbl:ch4-dataset}
\begin{tabular}{ll}
\toprule
Property & Value \\
\midrule
Subjects & 10 (intact) \\
Sessions & 2 per subject (a train, b test) \\
Electrodes & 16 (two rings of 8) \\
Sampling rate & 2 kHz (processed to 1 kHz) \\
Gestures & 7 grasps + rest \\
Window / stride & 300 / 150 ms \\
Train split & session a, repetitions not divisible by 5 \\
Within-session test & session a, held-out repetitions \\
Cross-session test & all repetitions of session b \\
Primary metric & trial-voted macro-F1 \\
\bottomrule
\end{tabular}
\end{subtable}

\vspace{4mm}
\begin{subtable}{\textwidth}\centering
\caption{}\label{tbl:ch4-hyperparams}
\begin{tabular}{ll}
\toprule
Hyper-parameter & Value \\
\midrule
Model width $d$ & 256 \\
Transformer layers / heads & 4 / 4 \\
Tokenizer channels & 32, 64, 128 \\
Parameters & 3.3M \\
Optimiser & AdamW, weight decay 0.01 \\
Learning rate & $3\times10^{-4}$, cosine decay \\
Epochs & 35 \\
Channel masking (train) & bounded, up to 40\% \\
Alignment momentum & target-batch statistics \\
\bottomrule
\end{tabular}
\end{subtable}
\end{table}

We employed NinaPro DB6~\citep{palermo2017repeatability} as our benchmark dataset; see
Table~\ref{tbl:ch4-dataset} for details. Ten intact participants completed 7 grip types + Rest
across sessions recorded on different days with 16 electrodes in 2 rings around their forearm at
2000 Hz. All signals were band-pass/notch filtered; normalized by amplitude per recording;
resampled to 1000 Hz; and segmented into 300 ms windows with 150 ms strides between segments. To
avoid having too many rest samples dominate the class balance, we subsampled rest windows.

We compare our results to two baselines. One is an implementation of the Hudgins time-domain
feature set~\citep{hudgins1993new}: MAV + WL + ZC + SSC + RMS with LDA per user, fit on Session
A data only. The second baseline consists of two recent studies using CNN/T-CNN
architectures~\citep{zanghieri2020temponet,lin2023longterm} without any form of adaptation using
data from Session B. The results from these two baselines are read in per-window accuracy so
that they sit alongside our results. Our hyperparameter settings are provided in
Table~\ref{tbl:ch4-hyperparams}.


\section{Results}

\subsection{Cross-session performance}

\begin{table}[p]\centering
\caption{Cross-session recognition on DB6. (a) DB6 within- and cross-session recognition.
Macro-F1 is trial-voted; per-window accuracy is the frame-level metric used by the published
baselines. Values are mean over ten subjects. (b) Per-subject DB6 inter-session macro-F1:
encoder within-session, encoder cross-session with and without label-free alignment, and the
LDA cross-session baseline. The encoder exceeds the LDA baseline for nine of the ten
subjects; subject 6 is the sole exception. (c) Paired significance on DB6 cross-session
macro-F1. The encoder's margin over the LDA pipeline and the alignment gain are both
significant across the ten subjects.}\label{tbl:ch4-crosssession}
\begin{subtable}{\textwidth}\centering
\caption{}\label{tbl:ch4-headline}
{\small\setlength{\tabcolsep}{4pt}
\begin{tabular}{llccc}
\toprule
Model & Regime & Macro-F1 & Window acc. & $\Delta$F1 vs class. \\
\midrule
Proposed encoder & Within-session & 0.978 & 0.847 & +0.117 \\
Proposed encoder & Cross-session & 0.688 & 0.550 & +0.148 \\
\midrule
Hudgins + LDA & Within-session & 0.861 & 0.630 & -- \\
Hudgins + LDA & Cross-session & 0.540 & 0.437 & -- \\
\bottomrule
\end{tabular}}
\end{subtable}

\vspace{4mm}
\begin{subtable}{\textwidth}\centering
\caption{}\label{tbl:ch4-persubject}
{\small\setlength{\tabcolsep}{4pt}
\begin{tabular}{lccccc}
\toprule
Subject & Enc. within & Enc. cross & Enc. +align & LDA cross & Enc.$-$LDA \\
\midrule
s1 & 0.969 & 0.907 & 0.917 & 0.714 & +0.193 \\
s2 & 1.000 & 0.780 & 0.803 & 0.446 & +0.334 \\
s3 & 1.000 & 0.563 & 0.591 & 0.552 & +0.012 \\
s4 & 0.949 & 0.717 & 0.769 & 0.642 & +0.074 \\
s5 & 1.000 & 0.871 & 0.878 & 0.704 & +0.167 \\
s6 & 0.990 & 0.598 & 0.609 & 0.652 & -0.055 \\
s7 & 0.980 & 0.691 & 0.697 & 0.506 & +0.186 \\
s8 & 0.970 & 0.554 & 0.635 & 0.360 & +0.195 \\
s9 & 0.941 & 0.446 & 0.512 & 0.295 & +0.150 \\
s10 & 0.980 & 0.753 & 0.757 & 0.528 & +0.225 \\
\midrule
Mean & 0.978 & 0.688 & 0.717 & 0.540 & +0.148 \\
\bottomrule
\end{tabular}}
\end{subtable}

\vspace{4mm}
\begin{subtable}{\textwidth}\centering
\caption{}\label{tbl:ch4-significance}
{\small\setlength{\tabcolsep}{4pt}
\begin{tabular}{lcccc}
\toprule
Comparison & Mean $\Delta$ & 95\% CI & Wilcoxon $p$ & Paired-$t$ $p$ \\
\midrule
Encoder $-$ LDA & +0.148 & [+0.081, +0.212] & 5.9e-03 & 2.3e-03 \\
Alignment $-$ no-adapt & +0.029 & -- & 2.0e-03 & -- \\
\bottomrule
\end{tabular}}
\end{subtable}
\end{table}

\begin{figure}[tbp]\centering
\begin{subfigure}{0.52\textwidth}\centering
\includegraphics[width=\linewidth]{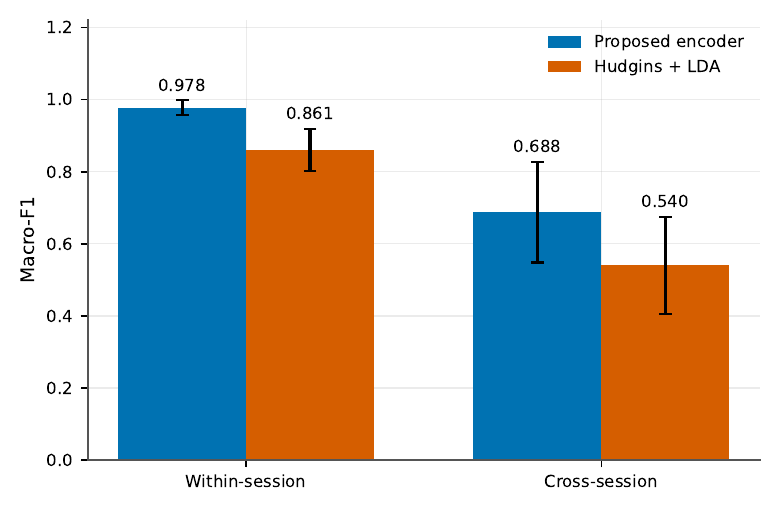}
\caption{}\label{fig:ch4-robustness}
\end{subfigure}\hfill
\begin{subfigure}{0.46\textwidth}\centering
\includegraphics[width=\linewidth]{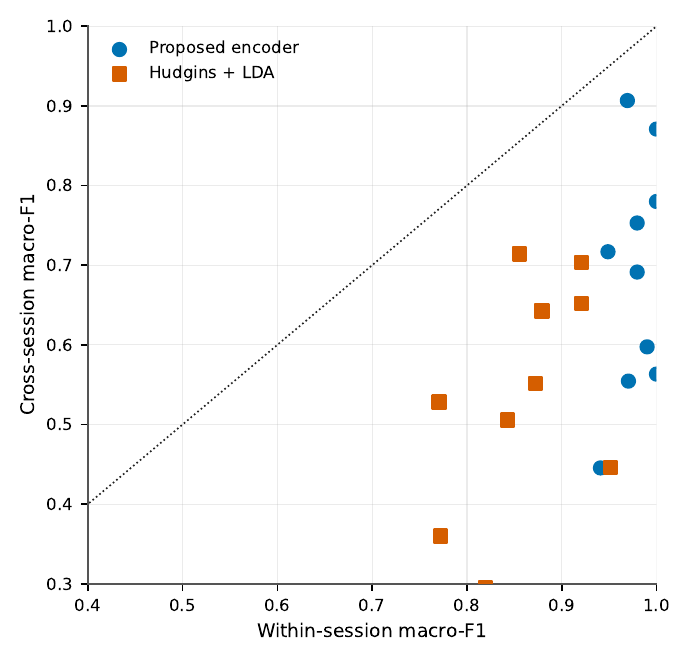}
\caption{}\label{fig:ch4-scatter}
\end{subfigure}

\vspace{3mm}
\begin{subfigure}{0.76\textwidth}\centering
\includegraphics[width=\linewidth]{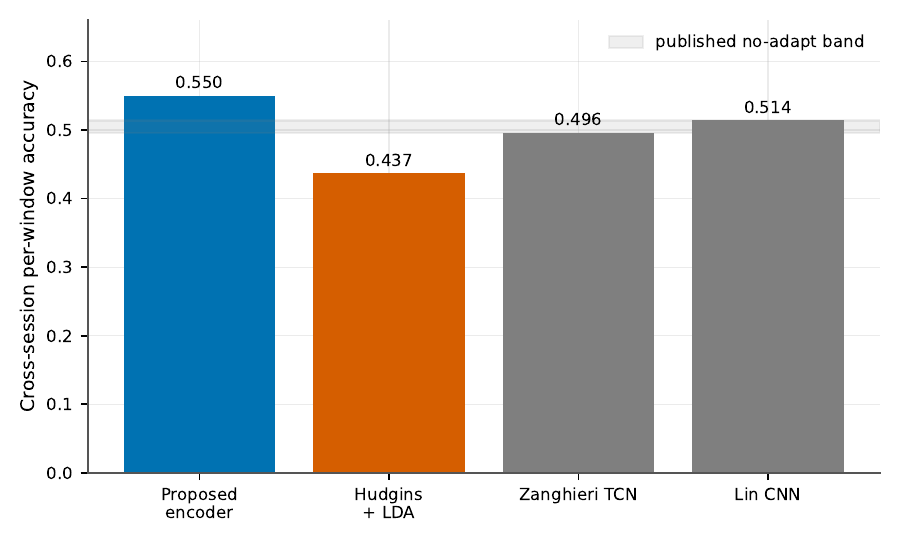}
\caption{}\label{fig:ch4-matched}
\end{subfigure}
\caption{Encoder against the baselines on DB6. (a) Session robustness. Within-session and
cross-session macro-F1 for the proposed encoder and the per-user Hudgins-and-LDA pipeline on
DB6. The encoder is ahead in both regimes and degrades less across the session change.
(b) Within versus cross-session macro-F1 per subject. The encoder points sit far above the
LDA cloud and closer to the identity line. (c) Like-for-like comparison. The encoder's
cross-session per-window accuracy sits in the upper part of the no-adaptation band reported
for DB6, above both published source-only anchors and well above the LDA pipeline.}
\label{fig:ch4-aggregate}
\end{figure}

As shown in Table~\ref{tbl:ch4-headline}, the encoder reached a score of 0.978 macro-F1 within a
session versus 0.861 for the LDA pipeline. When there is a day gap between the sessions, the
encoder drops to 0.688 while the LDA pipeline drops to 0.540. Because the encoder's
cross-session margin over the LDA pipeline is +0.148 macro-F1, larger than its within-session
margin, we can say the encoder lost less to the day gap than the LDA pipeline did. We also have
a figure showing these same four values as grouped bars (Figure~\ref{fig:ch4-robustness}) and
another showing each subject as a point in a scatter plot of within session vs. cross session
where the encoder has a much higher cloud above the LDA cloud and closer to the identity line
(Figure~\ref{fig:ch4-scatter}).

\begin{figure}[tbp]\centering
\begin{subfigure}{0.94\textwidth}\centering
\includegraphics[width=\linewidth]{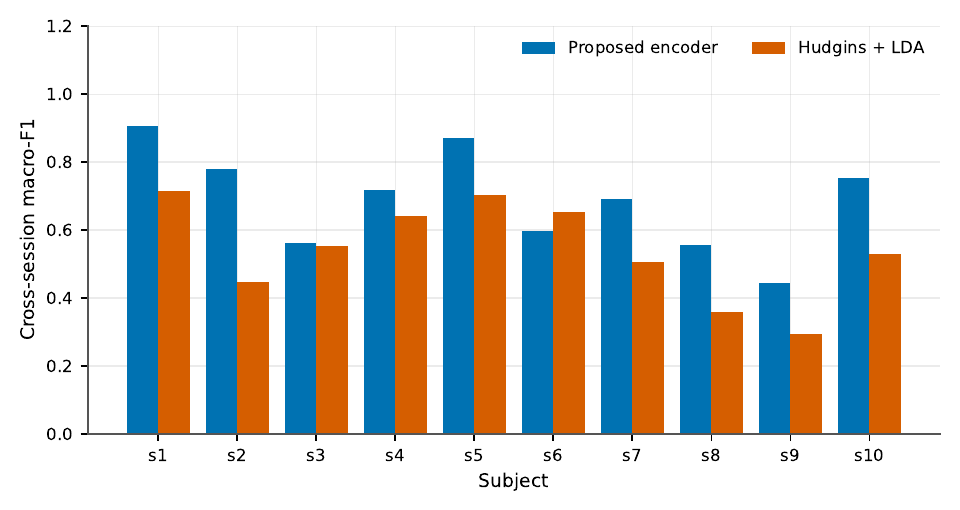}
\caption{}\label{fig:ch4-per_subject}
\end{subfigure}

\vspace{3mm}
\begin{subfigure}{0.94\textwidth}\centering
\includegraphics[width=\linewidth]{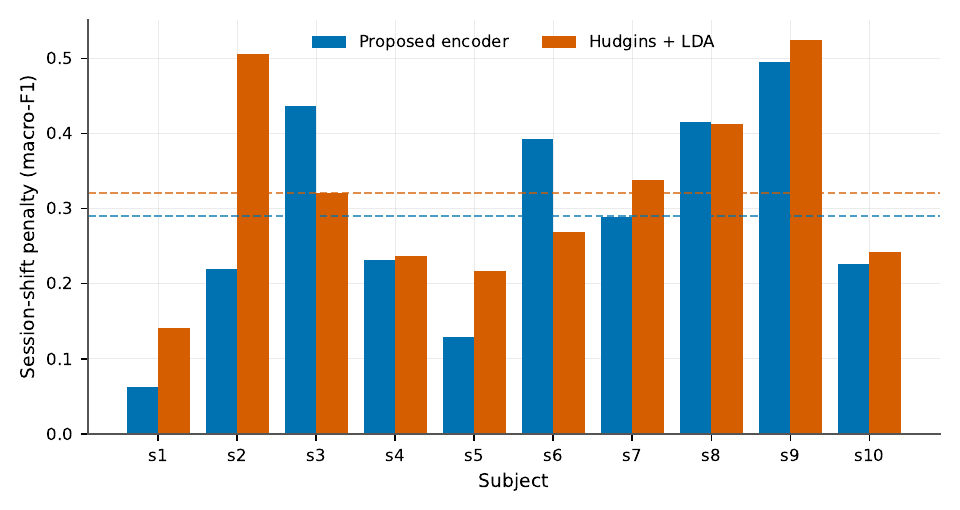}
\caption{}\label{fig:ch4-drop}
\end{subfigure}
\caption{Per-subject breakdown on DB6. (a) Per-subject cross-session macro-F1 on DB6. Nine of
the ten subjects exceed the LDA cross-session baseline, two of them by more than twenty
points; subject 6 is the sole exception. (b) Per-subject session-shift penalty (within minus
cross-session macro-F1). The encoder degrades less than the LDA pipeline for seven of the ten
subjects.}
\label{fig:ch4-persubject-fig}
\end{figure}

Also in Table~\ref{tbl:ch4-persubject} and Figure~\ref{fig:ch4-per_subject}, we break down the
mean cross session score for each subject. All but subject 6 have a better cross session score
with the encoder, two of them by more than twenty points. Subject 6 had a better cross session
score with the LDA pipeline (Encoder = 0.598; LDA = 0.652). We also calculate the session-shift
penalty, the within minus cross-session drop per subject, in Figure~\ref{fig:ch4-drop}. Seven of
the ten subjects experience less of a decline using the encoder than they do with the LDA
pipeline. Finally, we present paired tests in Table~\ref{tbl:ch4-significance}; we compute these
tests across all ten subjects. These tests indicate that our encoder is significantly better
than the LDA pipeline in cross session macro-F1 under both a Wilcoxon signed rank test and a
paired $t$-test. Also, we generate a bootstrap confidence interval around the mean difference
between our encoder and the LDA pipeline. This confidence interval excludes zero, indicating
that our encoder performs better than the LDA pipeline in cross session macro F1.

\subsection{Comparison with published DB6 results}

\begin{table}[t]\centering
\caption{Cross-session DB6 in matched, per-window accuracy against published source-only (no test-time adaptation) results on the same dataset. The proposed encoder sits at the top of the reported band.}\label{tbl:ch4-matched}
\begin{tabular}{lcl}
\toprule
Method & Window acc. & Note \\
\midrule
\textbf{Proposed encoder} & \textbf{0.550} & this work, source-only \\
Zanghieri TCN & 0.496 & published, source-only \\
Lin CNN & 0.514 & published, source-only \\
Hudgins + LDA & 0.437 & this work, per-user \\
\bottomrule
\end{tabular}
\end{table}

Our cross session macro-F1 of 0.688 is very low against DB6 results in the eighties reported in
the literature. However, those results are not directly comparable. First, they either used
target data during training or during some form of adaptation. Second, they reported per window
accuracy versus trial voted macro-F1. Third, they included the rest class in their average.
Fourth, they restricted their scoring to only steady-state windows. When put on an equal footing
(i.e., source only per window accuracy with no target data) we find that our encoder actually
beats both published results (Table~\ref{tbl:ch4-matched} and Figure~\ref{fig:ch4-matched}).
Specifically, if we compare the encoder's per window accuracy of 0.550 across sessions to that
of temporal convolutional models at 0.496~\citep{zanghieri2020temponet} and CNNs at
0.514~\citep{lin2023longterm}, we see that it performs better than both. Furthermore, it also
performs better than the LDA pipeline, which scored 0.437. Since these results were reported by
their authors and we ran no experiments to replicate them here, this two-point band locates the
encoder rather than establishing a ranking.


\subsection{Label-free adaptation methods}

\begin{table}[tbp]\centering
\caption{Adaptation results. (a) Label-free test-time adaptation on DB6, each method scored
against its own no-adaptation baseline over ten subjects. Feature alignment is the only
method that improves every subject. (b) Batch-normalisation statistic re-estimation (AdaBN)
on DB6. Neither placement recovers the session gap; re-estimating tokenizer statistics over
channel-mixed windows collapses the model.}\label{tbl:ch4-adaptation}
\begin{subtable}{\textwidth}\centering
\caption{}\label{tbl:ch4-tta}
{\small\setlength{\tabcolsep}{4pt}%
\begin{tabularx}{\textwidth}{@{}l >{\raggedright\arraybackslash}X cccl@{}}
\toprule
Method & Mechanism & No-adapt & Adapted & $\Delta$ & Subj.+ \\
\midrule
feature-align & Align target embedding mean/variance to source & 0.688 & 0.717 & +0.029 & 10/10 \\
strong & Alignment + class-conditional matching + self-training & 0.691 & 0.713 & +0.022 & 9/10 \\
TENT+div & Entropy minimisation with a marginal-entropy regulariser & 0.680 & 0.696 & +0.015 & 8/10 \\
AdaBN-1 & Re-estimate one embedding BatchNorm on the target & 0.693 & 0.686 & -0.006 & 5/10 \\
pseudo-label & Confidence-filtered pseudo-label self-training & 0.688 & 0.675 & -0.012 & 2/10 \\
AdaBN-full & Re-estimate all tokenizer BatchNorm on the target & 0.676 & 0.034 & -0.641 & 0/10 \\
\bottomrule
\end{tabularx}}
\end{subtable}

\vspace{4mm}
\begin{subtable}{\textwidth}\centering
\caption{}\label{tbl:ch4-adabn}
{\small\setlength{\tabcolsep}{4pt}%
\begin{tabularx}{\textwidth}{@{}l >{\raggedright\arraybackslash}X ccc@{}}
\toprule
Variant & Placement & No-adapt & After AdaBN & $\Delta$ \\
\midrule
Embedding BatchNorm & single BN at the pooled embedding & 0.693 & 0.686 & -0.006 \\
Tokenizer BatchNorm & BatchNorm in every tokenizer conv & 0.676 & 0.034 & -0.641 \\
\bottomrule
\end{tabularx}}
\end{subtable}
\end{table}

\begin{figure}[t]\centering
\includegraphics[width=0.98\textwidth]{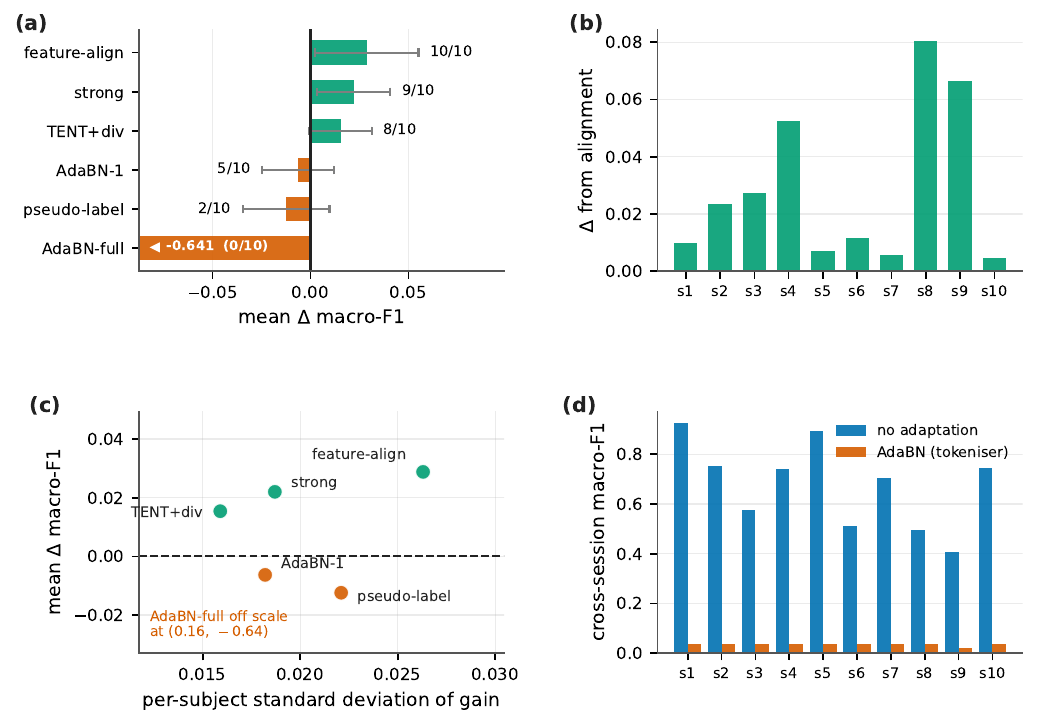}
\caption{Label-free adaptation on DB6. (a) Mean change in macro-F1 for each strategy against its own no-adaptation baseline, with the number of subjects improved; the axis is clipped to the comparable range and the batch-normalisation collapse is labelled where its bar leaves the frame. (b) Per-subject gain from feature alignment, positive for all ten. (c) Mean gain against per-subject spread, zoomed to the comparable methods. (d) Batch-normalisation re-estimation throughout the tokeniser, per subject: every subject collapses to a single class.}\label{fig:ch4-adaptation-panel}
\end{figure}

When there are no target labels available to train on in the new session, each of the five
different label-free adaptation methods behaves differently (Table~\ref{tbl:ch4-tta},
Figure~\ref{fig:ch4-adaptation-panel}a). The only method that improves every subject is feature
alignment. As shown in Figure~\ref{fig:ch4-adaptation-panel}b, it produces an average
improvement of +0.029 macro-F1. This represents improvements ranging from fractions of a point
to about eight points per subject. Entropy minimisation produces an improvement of +0.015 once
the anti-collapse regulariser is added; otherwise it collapses completely. By contrast, pseudo
labelling produces a decline of 0.012 in score, primarily because it uses target predictions
that are only about 67\% accurate as training targets, so that the errors reinforce themselves.
Only feature alignment appears in the reliable quadrant in
Figure~\ref{fig:ch4-adaptation-panel}c; it has a positive gain (+0.029) and a low variance.

\subsection{Batch-normalisation re-estimation}


Batch-normalisation re-estimation is treated separately because it is a standard label-free
method in this family~\citep{li2018adabn} (Table~\ref{tbl:ch4-adabn},
Figure~\ref{fig:ch4-adaptation-panel}d). Normalising a single layer at the pooled embedding
moves the score by -0.006. Normalising through all of the per-channel layers in the tokeniser
reduces the cross-session score to 0.034, effectively forcing all subjects into a single
classification. The cause is architectural; since each layer of the tokeniser processes a single
electrode at a time, a target batch will contain activations from all sixteen channels mixed
together into a common normalisation statistic. Therefore, re-estimating normalisation
statistics will combine channel-by-channel activation scales that were never intended to be
combined under a single normaliser; as a result, the pooled representation will degrade. There
is thus an architectural compatibility problem: the methods commonly employed in the domain
adaptation literature assume that normalisation layers operate on consistent feature semantics;
by design, this type of tokeniser does not provide this consistency.

\subsection{Label-free alignment against calibration}

\begin{table}[tbp]\centering
\caption{Calibration budgets and the strategy summary. (a) Target-session calibration
efficiency on DB6. Each budget's gain is measured against its own no-adaptation baseline. A
single labelled repetition matches the best label-free method; a third adds almost nothing.
(b) Summary of adaptation strategies for the cross-session setting, ordered by reliability.
Only feature alignment improves every subject without any target
labels.}\label{tbl:ch4-calibration}
\begin{subtable}{\textwidth}\centering
\caption{}\label{tbl:ch4-kshot}
{\small\setlength{\tabcolsep}{4pt}
\begin{tabular}{lccc}
\toprule
Budget & Approx. minutes & Cross-session macro-F1 & $\Delta$ vs 0-shot \\
\midrule
0-shot & 0.0 & 0.710 & +0.000 \\
1-shot & 0.4 & 0.738 & +0.029 \\
3-shot & 1.2 & 0.739 & +0.028 \\
\midrule
Label-free alignment & 0.0 & 0.717 & +0.029 \\
\bottomrule
\end{tabular}}
\end{subtable}

\vspace{4mm}
\begin{subtable}{\textwidth}\centering
\caption{}\label{tbl:ch4-summary}
{\small\setlength{\tabcolsep}{4pt}
\begin{tabular}{llcl}
\toprule
Strategy & Labels & Mean $\Delta$F1 & Outcome \\
\midrule
feature-align & none & +0.029 & Reliable (all subjects positive) \\
strong & none & +0.022 & Reliable but not better than alignment \\
TENT+div & none & +0.015 & Marginal, needs anti-collapse term \\
AdaBN-1 & none & -0.006 & Neutral to harmful \\
pseudo-label & none & -0.012 & Harmful (error reinforcement) \\
AdaBN-full & none & -0.641 & Collapses to one class \\
\midrule
1-shot calibration & 1 rep & +0.029 & Best small-label option \\
3-shot calibration & 3 reps & +0.028 & Plateaus early \\
\bottomrule
\end{tabular}}
\end{subtable}
\end{table}

\begin{figure}[tbp]\centering
\begin{subfigure}{0.52\textwidth}\centering
\includegraphics[width=\linewidth]{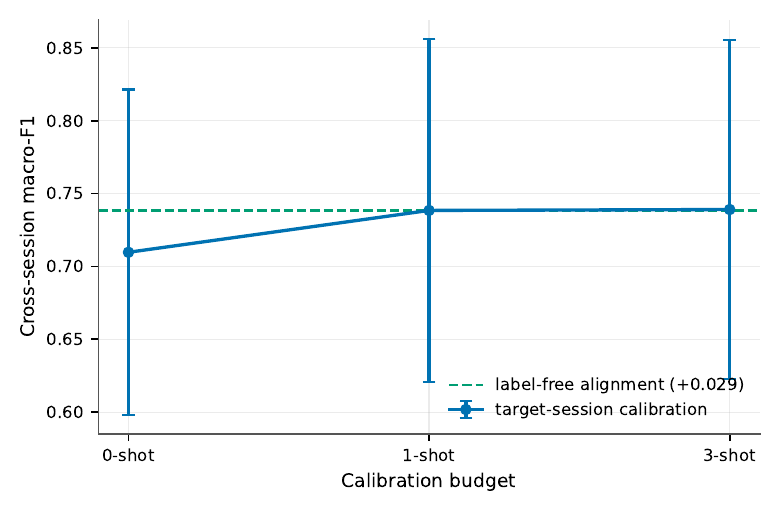}
\caption{}\label{fig:ch4-kshot}
\end{subfigure}\hfill
\begin{subfigure}{0.46\textwidth}\centering
\includegraphics[width=\linewidth]{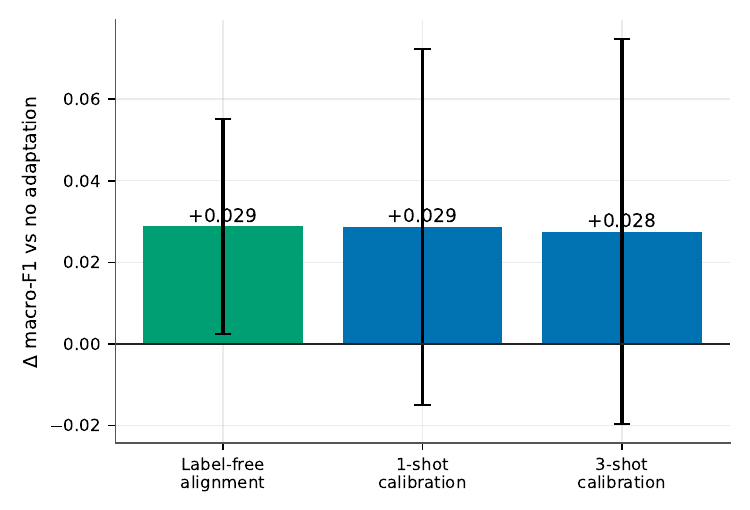}
\caption{}\label{fig:ch4-calib_free}
\end{subfigure}

\vspace{3mm}
\begin{subfigure}{0.68\textwidth}\centering
\includegraphics[width=\linewidth]{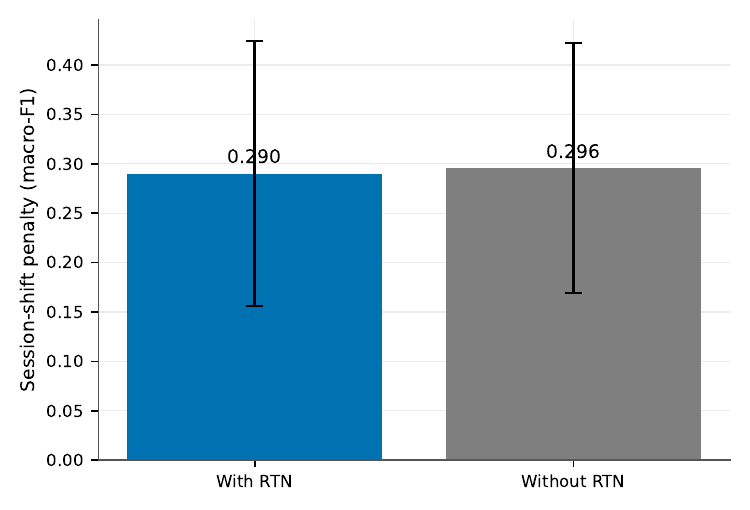}
\caption{}\label{fig:ch4-rtn}
\end{subfigure}
\caption{Closing the residual gap. (a) Calibration-efficiency on the target session. A single
labelled repetition recovers about the same gain as the best label-free method, and a third
repetition adds almost nothing. (b) Label-free alignment against light calibration. Feature
alignment reaches the one-shot calibration gain without any target labels. (c) Rolling-time
normalisation and the session gap. Removing it barely changes the cross-session drop, so the
calibration-free normaliser that carries cross-user transfer is not what protects against
session drift.}
\label{fig:ch4-closing}
\end{figure}

Adding one labeled repetition on the target session increases the cross-session macro-F1 by
+0.029; adding a third repetition yields virtually no additional benefit
(Table~\ref{tbl:ch4-kshot}, Figure~\ref{fig:ch4-kshot}). In addition,
Figure~\ref{fig:ch4-calib_free} juxtaposes these two paths: label-free feature alignment obtains
virtually all of the benefits of one shot calibration without requiring any target labels at
all.

\subsection{Source of the session robustness}

\begin{table}[t]\centering
\caption{Rolling-time normalisation and session drift on DB6. Removing the calibration-free normaliser barely changes the cross-session penalty, so the component that drives cross-user transfer does not explain session robustness.}\label{tbl:ch4-ablation}
\begin{tabular}{lccc}
\toprule
Variant & Cross-session F1 & Session drop & $\Delta$ drop \\
\midrule
With RTN & 0.688 & 0.290 & -- \\
Without RTN & 0.676 & 0.296 & +0.006 \\
\bottomrule
\end{tabular}
\end{table}

The calibration free rolling-time normalizer carried forward from the cross-user
study~\citep{odeyemi_encoder} provides the cross-user transferability reported there, which
makes it the obvious candidate for the session robustness as well. An ablation based on random
number seed matching eliminates rolling-time normalization as a contributing factor. Eliminating
rolling-time normalization changed the cross-session penalty by only +0.006, i.e., from a 0.290
drop to a 0.296 drop (Table~\ref{tbl:ch4-ablation}, Figure~\ref{fig:ch4-rtn}). Therefore,
normalizing per-user amplitude differences is close to neutral against session-to-session drift,
so session robustness can be attributed to the encoder as a whole rather than to that single
component. Table~\ref{tbl:ch4-summary} shows the full set of adaptation strategies, ordered by
reliability.


\section{Discussion}

Session robustness here is something that was never explicitly planned for in the architecture
of this model; nevertheless, despite being designed for cross-user, cross-montage
transferability it manages to retain most of its recognition accuracy even after moving to a new
day without needing any modifications made to it. Moreover, on a matched comparison basis our
encoder ends up inside a band of previously published DB6 results rather than below it. We note
that our apparent shortfall in comparison to numbers reported in the literature dissolves once
we stop comparing apples and oranges; specifically, once the metric, the class set, the window
selection and the use of target data are equalised, the uncorrected comparison is seen to
understate what a source-only model already achieves.

On adaptation, our findings are generally unfavorable towards existing approaches for test-time
adaptation. Specifically, most of the label-free methods currently popularized in the recent
test-time adaptation literature do not generalize as well as we expected when applied to our
particular experimental paradigm. The lone exception among label-free methods is feature
alignment; it worked and worked uniformly across all subjects. To some degree this is
unsurprising given its relatively weak assumptions regarding what happens when transitioning
between sessions (specifically that the change in session is roughly equivalent to an affine
transformation of what gets pooled prior to processing). Entropy minimization is able to perform
acceptably provided it guards against collapse; otherwise it fails. Similarly, pseudo labeling
and batch-normalisation re-estimation failed; although batch-normalisation re-estimation failed
in ways that are unique to an architecture such as ours and would likely not have been apparent
on conventional architectures. Therefore, researchers interested in performing similar
cross-session adaptation should consider employing statistical alignment as opposed to
batch-normalisation re-estimation (see Tables~\ref{tbl:ch4-tta} and~\ref{tbl:ch4-adabn}).

There exists a residual session gap that was unaddressed by any of the label-free methods
studied here; furthermore one labeled repetition produced a slightly superior cross-session
macro-F1 value compared to any label-free method examined. This leaves us with a limited scope
as we only tested our encoder on ten intact subjects and therefore these results refer only to
session drift. Within these limits, our encoder demonstrated session robustness and feature alignment
was the label-free adaptation that held up.

\section{Conclusion}

Without having made any changes whatsoever to its architecture, our montage-agnostic encoder
retained most of its recognition accuracy when recordings taken at one session were replaced by
recordings taken at a second session; additionally when compared on a like-for-like basis to
most published DB6 baselines our encoder performed equally well or better, and sat above both
published source-only results on that benchmark, a band of two points that locates the encoder
rather than ranking it. Among label-free adaptations, aligning the encoder's feature statistics
to the new session is the only intervention that helps every subject, and it recovers about what
a single labelled calibration repetition would.

\section*{CRediT authorship contribution statement}
\textbf{Jethro Odeyemi:} Conceptualization, Methodology, Software, Formal analysis,
Investigation, Data curation, Visualization, Writing -- original draft, Writing -- review
and editing.

\textbf{W.J. (Chris) Zhang:} Conceptualization, Methodology, Resources, Supervision,
Project administration, Writing -- review and editing.

\section*{Declaration of competing interest}
The authors declare no competing financial interests or personal relationships that could have
appeared to influence the work reported in this paper.

\section*{Ethics statement}
This work is a secondary computational analysis of the publicly available NinaPro DB6
database~\citep{palermo2017repeatability}. No new human data were collected, and no
participants were recruited, contacted or identified by the author for this study. Ethical
approval and informed consent for the original DB6 recordings were obtained by the
investigators who created that database.

\section*{Funding}
This research did not receive any specific grant from funding agencies in the public, commercial, or not-for-profit sectors.

\section*{Data availability}
This study uses only publicly available data. The NinaPro databases are available from the
NinaPro consortium.

\bibliographystyle{elsarticle-num}
\bibliography{refs}

\begin{thebibliography}{10}
\expandafter\ifx\csname url\endcsname\relax
  \def\url#1{\texttt{#1}}\fi
\expandafter\ifx\csname urlprefix\endcsname\relax\def\urlprefix{URL }\fi
\expandafter\ifx\csname href\endcsname\relax
  \def\href#1#2{#2} \def\path#1{#1}\fi

\bibitem{palermo2017repeatability}
F.~Palermo, M.~Cognolato, A.~Gijsberts, H.~M{\"u}ller, B.~Caputo, M.~Atzori,
  Repeatability of grasp recognition for robotic hand prosthesis control based
  on {sEMG} data, IEEE International Conference on Rehabilitation Robotics
  (ICORR) (2017) 1154--1159.

\bibitem{du2017surface}
Y.~Du, W.~Jin, W.~Wei, Y.~Hu, W.~Geng, Surface emg-based inter-session gesture
  recognition enhanced by deep domain adaptation, Sensors (2017).
\newblock \href {https://doi.org/10.3390/s17030458}
  {\path{doi:10.3390/s17030458}}.

\bibitem{zhai2017self}
X.~Zhai, B.~Jelfs, R.~H.~M. Chan, C.~Tin, Self-recalibrating surface emg
  pattern recognition for neuroprosthesis control based on convolutional neural
  network, Frontiers in Neuroscience (2017).
\newblock \href {https://doi.org/10.3389/fnins.2017.00379}
  {\path{doi:10.3389/fnins.2017.00379}}.

\bibitem{jiang2017feasibility}
S.~Jiang, B.~Lv, W.~Guo, C.~Zhang, H.~Wang, X.~Sheng, Feasibility of
  wrist-worn, real-time hand, and surface gesture recognition via semg and imu
  sensing, IEEE Transactions on Industrial Informatics (2017).
\newblock \href {https://doi.org/10.1109/tii.2017.2779814}
  {\path{doi:10.1109/tii.2017.2779814}}.

\bibitem{odeyemi_encoder}
J.~Odeyemi, W.~J. Zhang, A montage-agnostic encoder for calibration-light
  cross-user gesture recognition from surface electromyography, arXiv preprint
  arXiv:TODO-ARXIV-IDPreprint (2026).

\bibitem{hudgins1993new}
B.~Hudgins, P.~Parker, R.~N. Scott, A new strategy for multifunction
  myoelectric control, IEEE Transactions on Biomedical Engineering 40~(1)
  (1993) 82--94.

\bibitem{rehman2018multiday}
M.~Z.~u. Rehman, A.~Waris, S.~O. Gilani, M.~Jochumsen, I.~K. Niazi, M.~Jamil,
  Multiday emg-based classification of hand motions with deep learning
  techniques, Sensors (2018).
\newblock \href {https://doi.org/10.3390/s18082497}
  {\path{doi:10.3390/s18082497}}.

\bibitem{qi2019intelligent}
J.~Qi, G.~Jiang, G.~Li, Y.~Sun, B.~Tao, Intelligent human-computer interaction
  based on surface emg gesture recognition, IEEE Access (2019).
\newblock \href {https://doi.org/10.1109/access.2019.2914728}
  {\path{doi:10.1109/access.2019.2914728}}.

\bibitem{ganin2016domain}
Y.~Ganin, E.~Ustinova, H.~Ajakan, P.~Germain, H.~Larochelle, F.~Laviolette,
  M.~Marchand, V.~Lempitsky, Domain-adversarial training of neural networks,
  Journal of Machine Learning Research 17~(59) (2016) 1--35.

\bibitem{du2017intersession}
Y.~Du, W.~Jin, W.~Wei, Y.~Hu, W.~Geng, Surface {EMG}-based inter-session
  gesture recognition enhanced by deep domain adaptation, Sensors 17~(3) (2017)
  458.

\bibitem{ketyko2019domain}
I.~Ket{\'y}k{\'o}, F.~Kov{\'a}cs, K.~Z. Varga, Domain adaptation for
  s{EMG}-based gesture recognition with recurrent neural networks, in:
  International Joint Conference on Neural Networks (IJCNN), IEEE, 2019, pp.
  1--7.

\bibitem{sun2016deepcoral}
B.~Sun, K.~Saenko, Deep {CORAL}: Correlation alignment for deep domain
  adaptation, in: European Conference on Computer Vision (ECCV) Workshops,
  2016, pp. 443--450.

\bibitem{li2018adabn}
Y.~Li, N.~Wang, J.~Shi, X.~Hou, J.~Liu, Adaptive batch normalization for
  practical domain adaptation, Pattern Recognition 80 (2018) 109--117.

\bibitem{wang2021tent}
D.~Wang, E.~Shelhamer, S.~Liu, B.~Olshausen, T.~Darrell, Tent: Fully test-time
  adaptation by entropy minimization, in: International Conference on Learning
  Representations (ICLR), 2021.

\bibitem{lee2013pseudo}
D.-H. Lee, Pseudo-label: The simple and efficient semi-supervised learning
  method for deep neural networks, in: ICML Workshop on Challenges in
  Representation Learning, 2013.

\bibitem{zanghieri2020temponet}
M.~Zanghieri, S.~Benatti, A.~Burrello, V.~Kartsch, F.~Conti, L.~Benini, Robust
  real-time embedded {EMG} recognition framework using temporal convolutional
  networks on a multicore {IoT} processor, Vol.~14, 2020, pp. 244--256.

\bibitem{lin2023longterm}
C.~Lin, et~al., Long-term stability of surface {EMG} pattern recognition and
  the effect of inter-session variability, IEEE Transactions on Neural Systems
  and Rehabilitation Engineering (2023).

\bibitem{vaswani2017attention}
A.~Vaswani, N.~Shazeer, N.~Parmar, J.~Uszkoreit, L.~Jones, A.~N. Gomez,
  {\L}.~Kaiser, I.~Polosukhin, Attention is all you need, in: Advances in
  Neural Information Processing Systems (NeurIPS), 2017.

\end{thebibliography}

\end{document}